\documentclass[letterpaper]{article} 
\usepackage{aaai2026}  
\usepackage{times}  
\usepackage{helvet}  
\usepackage{courier}  
\usepackage[hyphens]{url}  
\usepackage{graphicx} 
\usepackage{natbib}  
\usepackage{caption} 
\usepackage{algorithm}
\usepackage{algorithmic}

\usepackage{newfloat}
\usepackage{listings}
\DeclareCaptionStyle{ruled}{labelfont=normalfont,labelsep=colon,strut=off} 
\floatstyle{ruled}
\newfloat{listing}{tb}{lst}{}
\floatname{listing}{Listing}
\usepackage{bm}
\usepackage{amssymb}
\usepackage{amsmath}
\usepackage{multirow}
\usepackage{pifont}

\title{Zero-Reference Joint Low-Light Enhancement and Deblurring via Visual Autoregressive Modeling with VLM-Derived Modulation}
\author{
    Wei Dong, Han Zhou\thanks{Corresponding author and project leader}, Junwei Lin, Jun Chen
}
\affiliations{
    McMaster University \ \{dongw22, zhouh115, lin523, chenjun\}@mcmaster.ca
}

\usepackage{bibentry}

\begin{document}

\maketitle

\begin{abstract}
Real-world dark images commonly exhibit not only low visibility and contrast but also complex noise and blur, posing significant restoration challenges. Existing methods often rely on paired data or fail to model dynamic illumination and blur characteristics, leading to poor generalization. To tackle this, we propose a generative framework based on visual autoregressive (VAR) modeling, guided by perceptual priors from the vision-language model (VLM). Specifically, to supply informative conditioning cues for VAR models, we deploy an adaptive curve estimation scheme to modulate the diverse illumination based on VLM-derived visibility scores.
In addition, we integrate dynamic and spatial-frequency-aware Rotary Positional Encodings (SF-RoPE) into VAR to enhance its ability to model structures degraded by blur. Furthermore, we propose a recursive phase-domain modulation strategy that mitigates blur-induced artifacts in the phase domain via bounded iterative refinement guided by VLM-assessed blur scores. Our framework is fully unsupervised and achieves state-of-the-art performance on benchmark datasets. 
\end{abstract}

\begin{links}
\link{Code}{https://github.com/LowLevelAI/VAR-LIDE}
\end{links}
\section{Introduction}
\label{sec_intro}

\begin{figure*}
\setlength{\abovecaptionskip}{1mm}
\setlength{\parskip}{0mm} 
\setlength{\baselineskip}{0mm} 
\centering
\begin{minipage}[c]{0.32\textwidth}
    \centering
    \includegraphics[width = 1\textwidth]{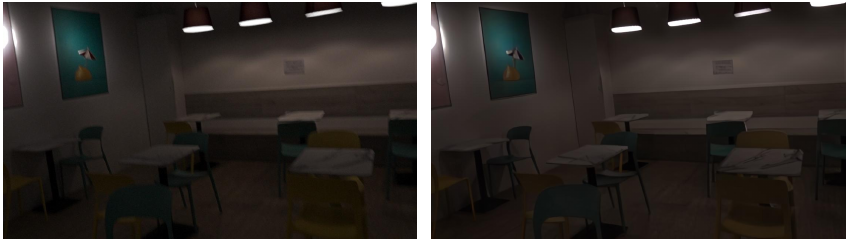}
    \setlength{\parskip}{0mm} 
    \setlength{\baselineskip}{0mm}
    \begin{minipage}[b]{0.48\linewidth}
    \centering
    \scriptsize{Noisy and Blurary Low-light Input}
    \end{minipage}
    \begin{minipage}[b]{0.48\linewidth}
    \centering
    \scriptsize{VAR \\ Output}
    \end{minipage}
    \includegraphics[width = 1\textwidth]{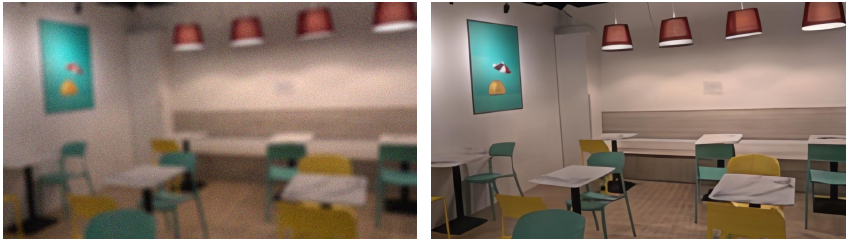}
    \setlength{\parskip}{0mm} 
    \setlength{\baselineskip}{0mm}
    \begin{minipage}[b]{0.48\linewidth}
    \centering
    \scriptsize{Noisy and Blurry Normal-light Input}
    \end{minipage}
    \begin{minipage}[b]{0.48\linewidth}
    \centering
    \scriptsize{VAR \\ Output}
    \end{minipage}
\caption*{ (a) Observations on blurry and noisy \\ inputs with pre-trained VAR models.}
\end{minipage}
\begin{minipage}[c]{0.645\textwidth}
    \centering
    \includegraphics[width = 1\textwidth]{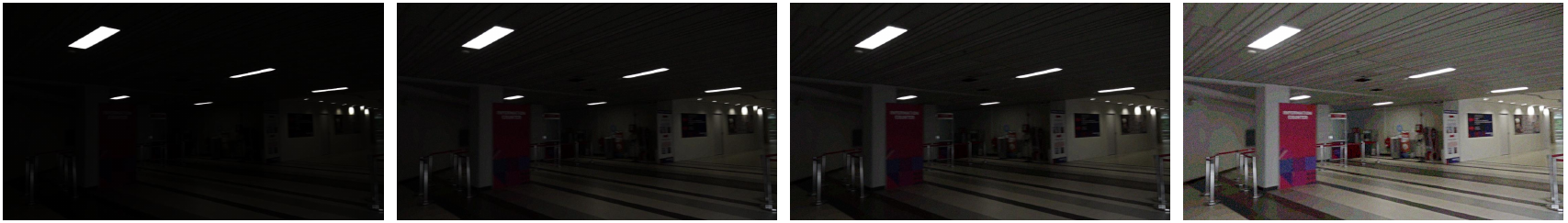}
    \setlength{\parskip}{0mm} 
    \setlength{\baselineskip}{0mm}
    \begin{minipage}[b]{0.24\linewidth}
    \centering
    \scriptsize{Extremely Dark Low-light Input}
    \end{minipage}
    \begin{minipage}[b]{0.24\linewidth}
    \centering
    \scriptsize{Adjustment Iteration $n = 4$ \textcolor{blue}{(\ding{55})}}
    \end{minipage}
    \begin{minipage}[b]{0.24\linewidth}
    \centering
    \scriptsize{Adjustment Iteration $n = 6$ \textcolor{blue}{(\ding{55})}}
    \end{minipage}
    \begin{minipage}[b]{0.24\linewidth}
    \centering
    \scriptsize{Adjustment Iteration $n = 8$ \textcolor{red}{($\checkmark$)}}
    \end{minipage}
    \includegraphics[width = 1\textwidth]{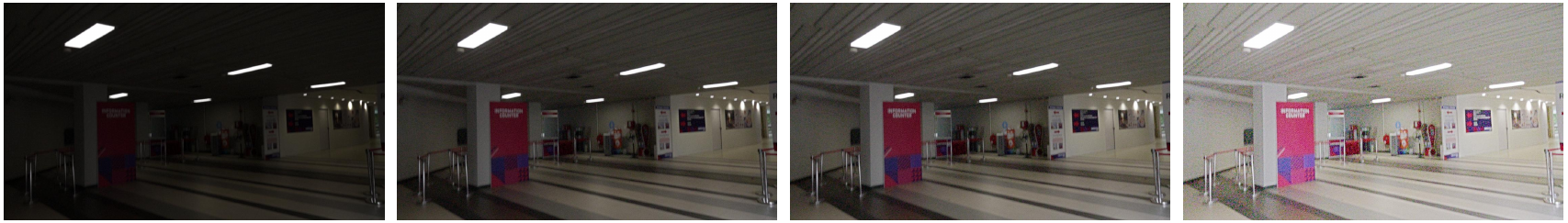}
    \setlength{\parskip}{0mm} 
    \setlength{\baselineskip}{0mm}
    \begin{minipage}[b]{0.24\linewidth}
    \centering
    \scriptsize{Low-light Input with Moderate Visibility}
    \end{minipage}
    \begin{minipage}[b]{0.24\linewidth}
    \centering
    \scriptsize{Adjustment Iteration $n = 4$ \textcolor{blue}{(\ding{55})}}
    \end{minipage}
    \begin{minipage}[b]{0.24\linewidth}
    \centering
    \scriptsize{Adjustment Iteration $n = 6$ \textcolor{red}{($\checkmark$)}}
    \end{minipage}
    \begin{minipage}[b]{0.24\linewidth}
    \centering
    \scriptsize{Adjustment Iteration $n = 8$ \textcolor{blue}{(\ding{55})}}
    \end{minipage}
\caption*{ (b) Zero-DCE presents unfavorable enhancement outcomes on low-light inputs with varying illumination conditions using a fixed adjustment iteration.}
\end{minipage}
\caption{Motivations of our proposed method. (a) We observe that the pre-trained VAR model~\cite{varsr} exhibits a certain degree of noise suppression and blur reduction, suggesting its potential for joint LLIE and deblurring. However, it struggles to substantially enhance visibility and recover fine structural details, underscoring the need for task-specific modules to better cope with real-world degradations. (b) Although the iterative illumination adjustment is generally effective, Zero-DCE~\cite{Zero-DCE} fails to provide satisfactory results under diverse illumination conditions, often causing under- or over-exposure artifacts due to its reliance on a fixed number of adjustment iterations.}
\label{fig_motivation}
\end{figure*}

The degradation of images captured in real-world dark environments can be formulated as: $\mathbf{x}_{LQ} = \gamma f(\mathbf{x}_{HQ}, \mathbf{k}) + \mathbf{n}$, with $\mathbf{x}_{HQ}$ denoting the high-quality (HQ) image and $\mathbf{x}_{LQ}$ its low-quality (LQ) counterpart. Here, $\mathbf{n}$ denotes sensor noise, $f$ is the convolution with the blur kernel $\mathbf{k}$ and $\gamma$ models dynamic range compression and saturation from exposure. Long exposure, a common strategy to improve photon capture, frequently leads to motion-induced blur and elevated noise levels. These combined artifacts degrade image quality and pose significant challenges for both human perception and high-level vision systems~\cite{sclera}.

Similar to other image restoration tasks~\cite{zhou2023breaking, dehazedct, dong2024shadowrefiner, rehit}, deep learning has led to notable progress in both low-light image enhancement (LLIE)~\cite{Retinexformer, enlightengan, Zero-DCE++, SCI} and deblurring~\cite{blur2blur, MambaIR, DWDN}, most methods treat them as separate tasks. LLIE models primarily boost brightness and reduce noise but often leave motion blur unaddressed. Conversely, deblurring algorithms typically assume sufficient illumination and perform poorly under low visibility. Although sequential pipelines may appear viable, they often disrupt blur cues during enhancement or fail to recover motion details when visibility is too low.

Although recent efforts explore joint LLIE and deblurring, most rely on supervised pipelines that require paired training data~\cite{DarkIR, LEDNet}, which is difficult to obtain in real-world scenarios. Though diffusion-based unsupervised methods~\cite{FourierDiff} show promising results, their reliance on lengthy sampling steps (\textit{e.g.}, 1,000 iterations) severely limits efficiency, making them unsuitable for practical applications.

Recently, visual autoregressive (VAR)~\cite{var} models offer a compelling alternative to diffusion methods by progressively generating high-resolution images via scale-wise token prediction, achieving superior structural fidelity and significantly faster inference without relying on costly iterative denoising. Technically, VAR models effectively preserve bidirectional spatial correlations while aligning with the unidirectional nature of autoregressive modeling, 
making them particularly suitable for image restoration tasks~\cite{varsr, VarFormer}, where LQ inputs serve as conditioning signals to guide the generation process. As illustrated in Fig.~\ref{fig_motivation}(a), our preliminary experiments indicate that pre-trained VAR models~\cite{varsr} possess inherent capabilities for noise suppression and partial blur reduction, making them a promising backbone for reference-free joint LLIE and deblurring. 

On the other hand, we observe that this VAR backbone demonstrates insufficient capacity for illumination enhancement and blur-compensated recovery, motivating the integration of specialized components to address visibility and detail degradation more effectively. An intuitive way to enable VAR with illumination correction capability is to incorporate a lightweight enhancement module prior to the generative process, allowing the model to operate on visibility-improved inputs while leveraging its inherent noise suppression properties. In our preliminary exploration, we adopt Zero-DCE~\cite{Zero-DCE}, a lightweight yet highly effective method for real-world LLIE, where deep neural networks are employed to predict the parameters of a differentiable curve-based model that iteratively adjusts image illumination. Once training converges, we observe that most pixels in the predicted curve parameter maps exhibit positive values, implying that increased iteration steps correspond to stronger illumination enhancement. However, this introduces two critical issues: (i) for moderately bright images, the default setting of 8 iterations tends to cause overexposure; (ii) reducing the iteration number alleviates overexposure but leads to insufficient enhancement on extremely dark images, as illustrated in Fig.~\ref{fig_motivation}(b). These observations indicate that a fixed iteration setting cannot robustly handle the diverse luminance conditions. This motivates further exploration into mechanisms that enable adaptive brightness modulation, ensuring consistent and perceptually compelling enhancement across diverse lighting scenarios.

Building upon these observations and insights, we introduce \textbf{VAR-LIDE}, a fully unsupervised generative framework for joint L\textbf{LI}E and \textbf{DE}blurring, which leverages the strengths of \textbf{VAR} models and perceptual guidance from the vision-language model (VLM). To effectively condition the VAR backbone, we develop a VLM-informed conditioning module that predicts adaptive enhancement curves based on VLM-assessed visibility, enabling robust performance under varying lighting conditions. Furthermore, we augment the pre-trained VAR backbone with dynamic and frequency-aware Rotary Positional Encodings to better model spatial structures degraded by motion blur. To mitigate motion-induced repeated edge artifacts in the Fourier phase domain, we introduce a recursive modulation mechanism that progressively refines the phase via a bounded parametric update, guided by blur-related VLM assessments. 

We summarize our contributions as following:

$\diamond$ We introduce \textbf{VAR-LIDE}, a fully unsupervised VAR-based framework that jointly addresses low-light image enhancement and deblurring, leveraging perceptual priors.

$\diamond$ Based on perceptual priors derived from VLM, we develop a VLM-informed conditioning module to support informative conditioning for VAR backbone, and design a recursive phase refinement mechanism to suppress blur-induced edge artifacts in the Fourier domain.

$\diamond$ We enhance the VAR backbone with content-aware spatial-frequency rotary positional encodings to better capture structural information under blur degradation.

$\diamond$ Our \textbf{VAR-LIDE} relaxes the reliance on paired supervision and achieves compelling performance on challenging real-world low-light benchmarks.

\section{Related Works}
\label{sec_rela}

\begin{figure*}[!t]
    \centering
    \setlength{\abovecaptionskip}{1.2mm}
    \centering
    \includegraphics[width=0.9\linewidth]{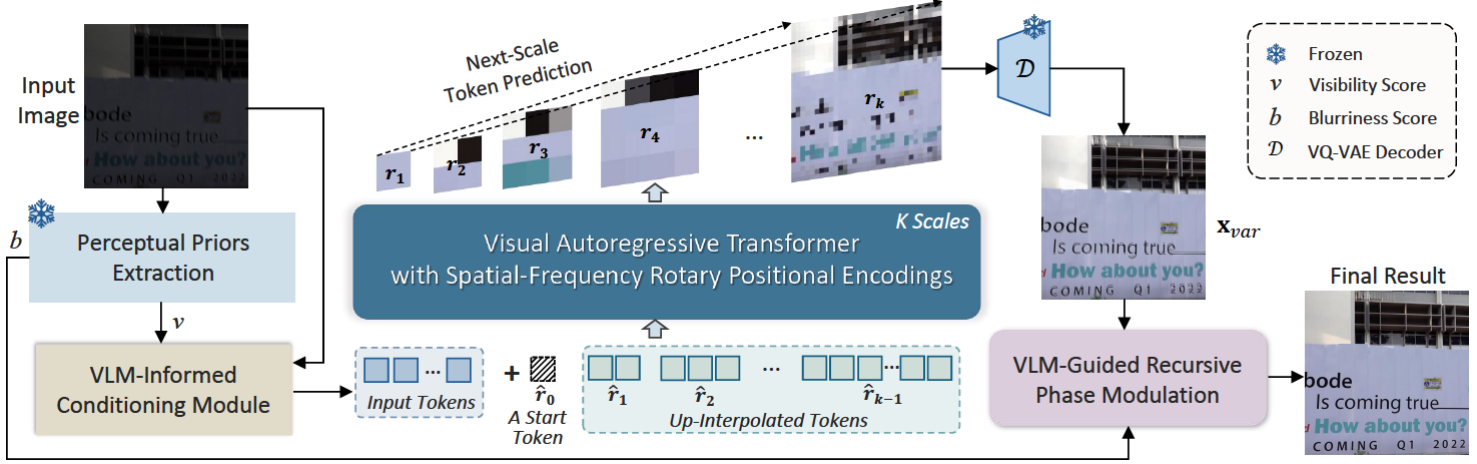}
    \caption{The overall framework of our proposed \textbf{VAR-LIDE} method, which adopts the pre-trained VAR model~\cite{varsr} as the backbone. We first leverage the perceptual priors extraction pipeline~\cite{gppllie} to acquire visibility-aware and blurriness-aware scores ($v$ and $b$). Then, $v$ is integrated into our VLM-Informed Conditioning Module (VICM) to adaptively improve the visibility and further support informative cues for VAR modeling. Moreover, to generate content-aware representations of positional embeddings, we develop the spatial-frequency rotary positional encodings (SF-RoPE) in VAR transformer blocks. Finally, guided by the VLM assessment $b$, we introduce a recursive modulation mechanism (VGPM) in the \texttt{FFT} phase domain to further mitigate blurriness and achieve visually compelling outputs.}
    \label{fig_framework}
\end{figure*}

\paragraph{LLIE and Deblurring as Separate Tasks} LLIE and deblurring are traditionally handled separately. Early LLIE methods \cite{HistogramEqualization, retinex, NaturalnessPreservedEnhancement} used hand-crafted priors, while recent deep models \cite{Zero-DCE, sgllie, glare, ecmamba, litags} learn brightness correction but neither are able to remove real-world blur, limiting their practical value. In parallel, traditional deblurring methods utilize predefined kernels for deconvolution \cite{BlindImageDeconvolution, RealisticDegradations}, while deep learning models \cite{DWDN, GRL, blur2blur} aim for better generalization. However, these methods assume well-lit inputs, which LLIE results may not meet, leading to artifacts and degraded performance.

\paragraph{Joint LLIE and Deblurring} Joint LLIE and deblurring has attracted increasing attention. Supervised methods~\cite{LEDNet, DarkIR} depend on costly paired data, whereas unsupervised approaches \cite{SSFlow, FourierDiff} use reconstruction or contrastive objectives to avoid this limitation. The architectures have evolved from CNNs~\cite{LEDNet} to transformers~\cite{Retinexformer}, Mamba~\cite{LIEDNet}, normalizing flow~\cite{SSFlow}, and diffusion models~\cite{FourierDiff}. However, achieving efficient and generalizable joint restoration remains challenging.

\paragraph{Visual Autoregressive Modeling} VQ-VAE \cite{vqvae} encodes images as quantized tokens for autoregressive generation but lacks spatial awareness. VAR \cite{var} improves this via next-scale token prediction, boosting quality and speed. Though applied to synthesis tasks \cite{CoDe, Infinity}, VAR is underexplored in image restoration under degradations like low-light blur. Recent work \cite{VarFormer, RestoreVAR, varsr} demonstrates its potential for image restoration by leveraging multiscale priors. We employ the VAR backbone for the joint LLIE and deblurring task, augmented with modules for illumination modulation and blur suppression.

\begin{figure}[!t]
    \centering
    \setlength{\abovecaptionskip}{1.2mm}
    \centering
    \includegraphics[width=0.85\linewidth]{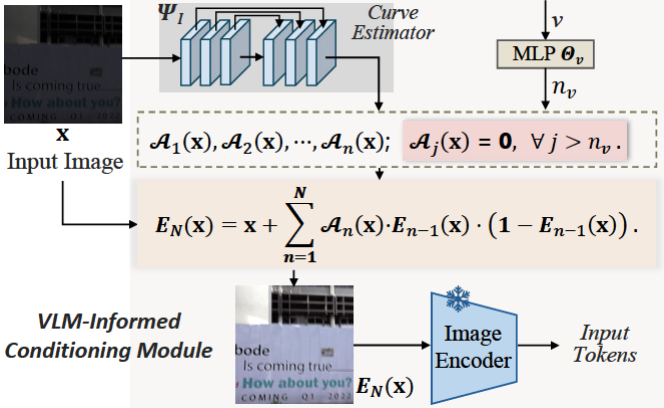}
    \caption{The overall framework of our VICM. It estimates illumination curves and adaptively truncates them based on a visibility-aware iteration count $n_v$.}
    \label{fig_vicm}
\end{figure}
\section{Method}
\label{sec_method}

\begin{figure*}
\setlength{\abovecaptionskip}{1mm}
\setlength{\parskip}{0mm} 
\setlength{\baselineskip}{0mm} 
\centering
\begin{minipage}[c]{1\textwidth}
    \centering
    \includegraphics[width = 1\textwidth]{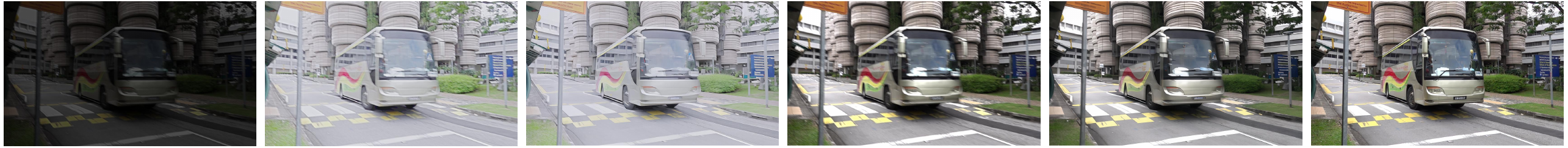}
    \setlength{\parskip}{0mm} 
    \setlength{\baselineskip}{0mm}
    \begin{minipage}[b]{0.16\linewidth}
    \centering
    \scriptsize{ (i) Low-quality \\ Input}
    \end{minipage}
    \begin{minipage}[b]{0.16\linewidth}
    \centering
    \scriptsize{ (ii) Zero-DCE \\ Enhanced Result}
    \end{minipage}
    \begin{minipage}[b]{0.16\linewidth}
    \centering
    \scriptsize{ (iii) VARSR Output with Condition (ii)}
    \end{minipage}
    \begin{minipage}[b]{0.16\linewidth}
    \centering
    \scriptsize{ (iv) Enhancement of \\ Our Proposed VICM}
    \end{minipage}
    \begin{minipage}[b]{0.16\linewidth}
    \centering
    \scriptsize{ (v) VARSR Output with Condition (iv)}
    \end{minipage}
    \begin{minipage}[b]{0.16\linewidth}
    \centering
    \scriptsize{ (vi) GT \\ Reference}
    \end{minipage}
\caption*{ (a) Our proposed VICM module provides more effective conditional cues for VAR, achieving more satisfactory results.}
\end{minipage}
\begin{minipage}[c]{1\textwidth}
    \centering
    \includegraphics[width = 1\textwidth]{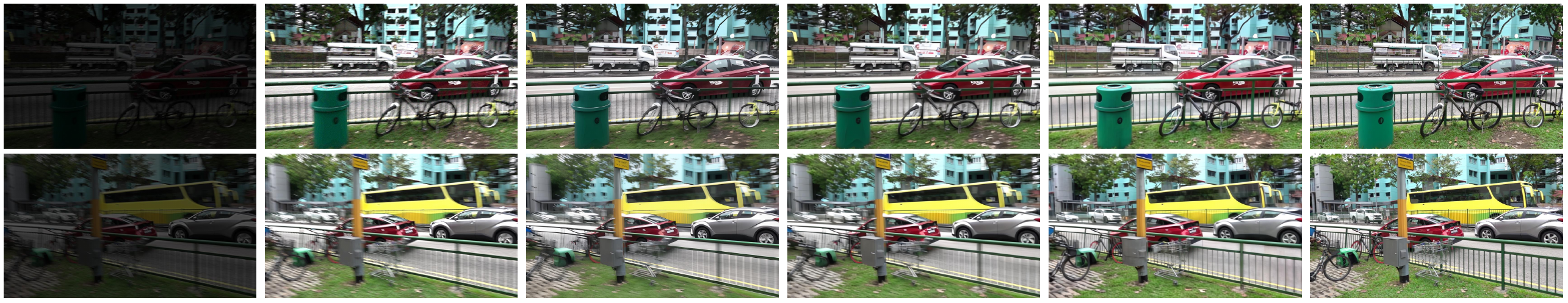}
    \setlength{\parskip}{0mm} 
    \setlength{\baselineskip}{0mm}
    \begin{minipage}[b]{0.16\linewidth}
    \centering
    \scriptsize{(i) Low-quality Input}
    \end{minipage}
    \begin{minipage}[b]{0.16\linewidth}
    \centering
    \scriptsize{(ii) w/ VICM only}
    \end{minipage}
    \begin{minipage}[b]{0.16\linewidth}
    \centering
    \scriptsize{ (iii) + Vanilla RoPE}
    \end{minipage}
    \begin{minipage}[b]{0.16\linewidth}
    \centering
    \scriptsize{ (iv) + Our SF-RoPE}
    \end{minipage}
    \begin{minipage}[b]{0.16\linewidth}
    \centering
    \scriptsize{(v) + VGPM (Full model)}
    \end{minipage}
    \begin{minipage}[b]{0.16\linewidth}
    \centering
    \scriptsize{(vi) GT}
    \end{minipage}
\caption*{ (b) Progressive Enhancement via VICM, SF-RoPE, and VGPM. Vanilla RoPE is adopted from VARSR~\cite{varsr}}
\end{minipage}
\caption{Comparative visualization of conditioning effects and enhancement quality across modules in our framework.}
\label{fig_observation_across}
\end{figure*}

The main focus of this work is to extend the capabilities of pre-trained Visual Autoregressive (VAR) models~\cite{var} to the challenging task of joint low-light image enhancement (LLIE) and deblurring. While the most relevant prior work, VARSR~\cite{varsr} explores scale-aligned rotary positional encodings (SA-RoPE) and diffusion refiners for image super-resolution, our method targets a different problem domain and proposes three novel modules tailored for real-world degradation: (i) a VLM-informed conditioning module for perceptual-aware illumination control (Sec.~\ref{sec_method_conditioning}), (ii) a spatial-frequency RoPE mechanism that fuses FFT-phase guided rotation with scale-aligned spatial encoding (Sec.~\ref{sec_method_rope}), and (iii) a recursive phase modulation module that explicitly targets blur-induced phase duplication artifacts (Sec.~\ref{sec_method_phase_blur}). Lastly, a reference-free optimization strategy is proposed to enable training without ground-truth supervision (Sec.~\ref{sec_method_optimization}). Our framework is illustrated in Fig.~\ref{fig_framework}.

\subsection{VLM-Informed Conditioning Module}
\label{sec_method_conditioning}

In VAR-based restoration~\cite{varsr}, the low-quality (LQ) input is embedded as prefix tokens to guide the generation process, making the informativeness of these conditional cues crucial for reconstruction fidelity. To strengthen the generative conditioning, we propose a VLM-Informed Conditioning Module (VICM) that adaptively modulates luminance based on perceptual cues.

Our design is motivated by the limitations of heuristic illumination adjustment strategies (\textit{e.g.}, Zero-DCE) in serving as effective conditioning for generative restoration models. Although Zero-DCE improves brightness, it lacks adaptability across diverse lighting conditions. As shown in Fig.~\ref{fig_motivation}(b), shallow enhancement (\textit{e.g.}, $n{=}4$) results in under-exposure in extremely dark scenes, while deeper enhancement (\textit{e.g.}, $n{=}8$) causes overexposure in moderately lit inputs. These suboptimal adjustments (\textit{e.g.}, Fig.~\ref{fig_observation_across}(a)(ii)) degrade the conditioning quality and propagate artifacts in the generative output (Fig.~\ref{fig_observation_across}(a)(iii)).
Formally, Zero-DCE models enhancement as an iterative curve-based transformation:
\begin{equation}
    \bm{E}_N(\mathbf{x}) = \mathbf{x} + \sum_{n=1}^{N} \bm{\mathcal{A}}_n(\mathbf{x}) \cdot \bm{E}_{n-1}(\mathbf{x}) \cdot \left(\mathbf{1} - \bm{E}_{n-1}(\mathbf{x}) \right),
 \label{curve}
 \end{equation}
where $\bm{E}_0(\mathbf{x}) = \mathbf{x}$, $\bm{\mathcal{A}}_n(\mathbf{x})$ denotes the curve parameter at iteration $n$, and $N$ is the total number of iterations. While this formulation captures nonlinear illumination trends, its fixed-step ($N$) nature fundamentally limits adaptiveness.

To address this limitation, we incorporate perceptual priors extracted from the vision-language model (VLM) following GPP-LLIE~\cite{gppllie}. A visibility-aware score $v$ is first computed via the \textit{Global Perceptual Prior Extraction Pipeline} in GPP-LLIE, which is then processed by a lightweight MLP $\boldsymbol{\Theta}_v$ to estimate an optimal iteration count $n_v$. As illustrated in Fig.~\ref{fig_framework}, this $n_v$ is used to truncate the illumination adjustment process within our VICM. Specifically, the curve estimator $\bm{\Psi}_I$ produces illumination curves $\{\bm{\mathcal{A}}_n(\mathbf{x}) \}_{n=1}^N$, and curve parameters beyond $n_v$ are masked to ensure perceptual adaptiveness:
\begin{equation}
\bm{\mathcal{A}}_j(\bm{\mathbf{x}}) = 0,\quad \forall j > n_v.
\end{equation}
This adaptive truncation ensures that illumination enhancement remains within a perceptually valid range. The enhanced image $\bm{E}_{N}(\mathbf{x})$ is then embedded and tokenized as a conditioning input to the VAR model. Compared with fixed-iteration enhancement pipelines, our VICM provides more informative and spatially adaptive guidance (Fig.~\ref{fig_observation_across}(a)(iv)), thereby improving downstream generation (Fig.~\ref{fig_observation_across}(a)(v)). Nonetheless, some structural artifacts remain (e.g., Fig.~\ref{fig_observation_across}(b)(iii)), motivating the design of complementary modules to better handle motion-related degradations.
\begin{figure*}[!t]
    \centering
    \setlength{\abovecaptionskip}{1.2mm}
    \centering
    \includegraphics[width=0.8\linewidth]{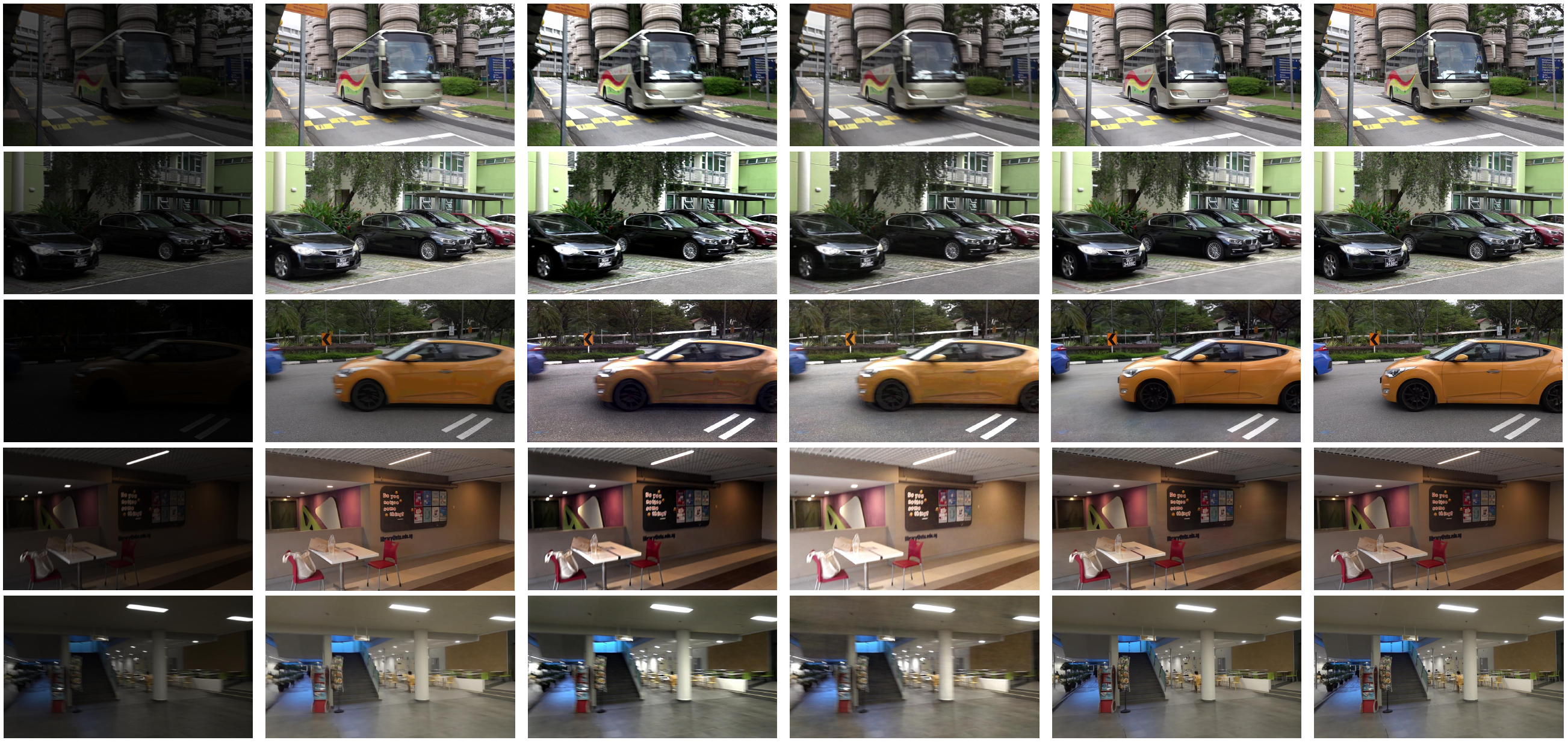}
    \setlength{\parskip}{0mm} 
    \setlength{\baselineskip}{0mm}
\begin{minipage}{0.8\linewidth}
    \begin{minipage}[b]{0.16\linewidth}
    \centering
    \scriptsize{ LQ}
    \end{minipage}
    \begin{minipage}[b]{0.16\linewidth}
    \centering
    \scriptsize{ Retinexformer}
    \end{minipage}
    \begin{minipage}[b]{0.16\linewidth}
    \centering
    \scriptsize{ SSFlow}
    \end{minipage}
    \begin{minipage}[b]{0.16\linewidth}
    \centering
    \scriptsize{ FourierDiff}
    \end{minipage}
    \begin{minipage}[b]{0.16\linewidth}
    \centering
    \scriptsize{\textbf{Ours}}
    \end{minipage}
    \begin{minipage}[b]{0.16\linewidth}
    \centering
    \scriptsize{Reference}
    \end{minipage}
\end{minipage}
    \caption{Visual comparisons on the LOL-Blur dataset, which involves both severe low-light conditions and motion blur. Compared to existing methods, our approach better preserves fine details and improves perceptual quality across diverse scenes.}
    \label{fig_lolblur}
\end{figure*}

\subsection{Content-Aware Spatial-Frequency RoPE}
\label{sec_method_rope}

To further alleviate structure-related artifacts observed in blur-degraded regions, we focus on enhancing the positional encoding mechanism within the VAR backbone. This refinement is motivated by the observation that existing rotary encoding schemes, such as those used in VARSR~\cite{varsr}, rely solely on position-indexed rotation matrices that lack sensitivity to content degradation. While Rotary Positional Encoding (RoPE) enables efficient modeling of relative positions by applying fixed sinusoidal rotations to query-key pairs, its static nature limits adaptability in structurally complex or spatially degraded regions.

To address this limitation, we propose a content-aware spatial-frequency RoPE (SF-RoPE) that modulates attention weights based on both positional and frequency-domain cues. Our approach introduces token-wise rotation matrices whose parameters are dynamically inferred from local frequency-phase statistics, thereby enabling finer control over attention in blur-sensitive areas.

\paragraph{Frequency-domain RoPE} At scale $K$, we extract frequency-phase information from the embedding $\mathbf{x}_{K-1}$ of token map $\mathbf{r}_{K-1}$ via \texttt{FFT}:
\begin{equation}
\begin{split}
    & \mathbf{F}(u, v) = \texttt{FFT}(\mathbf{x}_{K-1}), \Phi(u, v) =\arg(\mathbf{F}(u,v)), \\
    & \mathbf{R}_{\Phi(u, v)} =
    \begin{bmatrix}
        \cos(\Phi(u, v)) & -\sin(\Phi(u, v)) \\
        \sin(\Phi(u, v)) & \cos(\Phi(u, v))
    \end{bmatrix},
\end{split}
\end{equation}
where $(u, v)$ denotes the frequency coordinate and $\Phi(u,v)$ captures the local phase. We then construct a token-specific frequency-based RoPE matrix as:

\begin{equation}
\texttt{RoPE}_{\text{freq}}(\mathbf{x}_K^{(u,v)}) =
\begin{bmatrix}
\mathbf{R}_{\Phi(u, v)}^{\frac{C}{2}} & \mathbf{0}^{\frac{C}{2}} \\
\mathbf{0}^{\frac{C}{2}} & \mathbf{R}_{\Phi(u, v)}^{\frac{C}{2}}
\end{bmatrix},
\end{equation}
where $C$ denotes the channel dimension. This enables the attention mechanism to be directly modulated by local blur-sensitive frequency variations.

\paragraph{Scale-aligned RoPE} 
In parallel, we apply standard RoPE in spatial domain using scale-normalized token coordinates to ensure positional consistency across multiple resolutions. For each $\textbf{x}_k^{(i,j)}$, the spatial rotary matrix is computed as:
\begin{equation}
\texttt{RoPE}_{\text{spa}}(\textbf{x}_k^{(i,j)}) =
\begin{bmatrix}
\mathbf{R}_{\Theta,\left(\frac{ih_K}{h_k}\right)}^{\frac{C}{2}} & \mathbf{0}^{\frac{C}{2}} \\
\mathbf{0}^{\frac{C}{2}} & \mathbf{R}_{\Theta,\left(\frac{jw_K}{w_k}\right)}^{\frac{C}{2}}
\end{bmatrix},
\end{equation}
where $(i,j)$ denotes the spatial location, $(h_k, w_k)$ are the dimensions at scale $k$, and $(h_K, w_K)$ refer to the base scale.

\paragraph{Spatial-Frequency Fusion}
To adaptively leverage both structural and positional priors, we fuse frequency and spatial encodings using a learnable mixing coefficient $\lambda$:
\begin{equation}
\texttt{RoPE}_{\text{fused}} = 
\lambda \cdot \texttt{RoPE}_{\text{freq}}+ (1-\lambda) \cdot \texttt{RoPE}_{\text{spa}}.
\end{equation}
\noindent This fusion balances global positional alignment with local content sensitivity, allowing the attention module to better capture distortions induced by motion blur and low resolution. As illustrated in Fig.~\ref{fig_observation_across}(b)(iv), the incorporation of our SF-RoPE module yields sharper edge recovery and improved spatial coherence. Nevertheless, subtle distortions remain in highly cluttered areas (\textit{e.g.}, bicycle), motivating further refinement toward structural consistency.

\begin{table*}[t]
\Large
\setlength{\abovecaptionskip}{1mm}
\setlength{\tabcolsep}{1.3pt}
\renewcommand\arraystretch{1.0}
  \centering
  \resizebox{\linewidth}{!}{
  \begin{tabular}{cc|ccccc|cc|ccccc}
    \hline
    \hline
    \multirow{2}{*}{\begin{tabular}{c}
            \textbf{ Methods}
        \end{tabular}} &\multirow{2}{*}{\begin{tabular}{c}
            \textbf{ Type}
        \end{tabular}} &\multicolumn{5}{c|}{Metrics} &\multirow{2}{*}{\begin{tabular}{c}
            \textbf{ Methods}
        \end{tabular}} &\multirow{2}{*}{\begin{tabular}{c}
            \textbf{ Type}
        \end{tabular}} &\multicolumn{5}{c}{Metrics}\\
         & &PSNR$\uparrow$ &NIQE$\downarrow$ &LPIPS$\downarrow$ &FID$\downarrow$ &CLIPIQA$\uparrow$ & & &PSNR$\uparrow$ &NIQE$\downarrow$ &LPIPS$\downarrow$ &FID$\downarrow$ &CLIPIQA$\uparrow$ \\ 
    \hline

    \multicolumn{14}{c}{\textbf{Cascaded Methods}} \\ \hline
    EnlightenGAN + BD\_Noise  & L+D &17.25 &\textbf{4.98} &0.413 &47.66  &0.203 
    & BD\_Noise+ EnlightenGAN & D+L &17.11 &\textbf{4.96} &0.419 &\textbf{49.21}  &0.201  \\

     Zero-DCE++ + BD\_Noise & L+D &14.77 &5.65 &0.533 &57.45 &0.165 
     & BD\_Noise + Zero-DCE++ &D+L &14.71 &5.67 &0.537 &56.96 &0.164  \\

     SCI + BD\_Noise & L+D &14.34 &5.48 &0.541 &60.28 & 0.163
     &BD\_Noise + SCI &D+L &14.56 &5.37 &0.526 &57.44 & 0.165  \\

     EnlightenGAN + Blur2Blur & L+D &\textbf{18.16} &5.02 &\textbf{0.396} &\textbf{45.73} &\textbf{0.206} 
     &Blur2Blur +EnlightenGAN & D+L &\textbf{17.38} &4.98 &\textbf{0.395} &54.53 &\textbf{0.226}  \\

     Zero-DCE++ + Blur2Blur & L+D &15.79 &5.74 &0.529 &80.88 &0.198  
     & Blur2Blur + Zero-DCE++ &D+L &14.55 &5.85 &0.543 &60.06 &0.183 \\

     SCI + Blur2Blur  &L+D &16.10 &5.75 &0.507 &69.39 &0.169 
     &Blur2Blur+ SCI  &D+L &14.55 &5.43 &0.547 &68.05 &0.180   \\
     \hline

    \multicolumn{14}{c}{\textbf{Joint LLIE and Deblurring Methods}} \\ \hline

    LEDNet & w R &24.36  &5.37 & 0.227 & 25.19 & 0.207 
    &LIEDNet  & w R &26.25 &5.40 &0.133 &13.18 &0.292  \\

    Retinexformer  & w R &24.76  &6.07 &0.219 &22.58 &0.214 
     &LIEDNet-L  & w R &\textbf{26.42} &5.17 &\textbf{0.127} &\textbf{11.38} &\textbf{0.305}  \\
     
     JUDE  & w R &25.26  &5.87 &0.186 &22.11 & 0.247
     &SSFlow*  & w/o R &19.24 &5.93 &0.307 &42.05 &0.183   \\

    DarkIR-M  & w R &25.74 &5.28 &0.165 &16.35 &0.286 
     &FourierDiff  & w/o R & 20.22  &4.97 &0.441 &50.59 &0.161   \\

    DarkIR-L & w R &26.14 & \textbf{5.15} &0.146 &14.27 &0.291 
    & \textbf{Ours}  & w/o R &\textbf{23.39} &\textbf{4.80} &\textbf{0.191} &\textbf{26.04} &\textbf{0.262}   \\
     
  \hline
  \hline
  \end{tabular}
  }
  \caption{Quantitative comparisons on LOLBlur Dataset. `L+D' and `D+L' indicate that the methods belong to \textbf{LLIE}~$\rightarrow$~\textbf{Deblurring} and ~\textbf{Deblurring}~$\rightarrow$\textbf{LLIE} methods, respectively;  `w R' or `w/o R' represent that the methods are \textbf{Joint LLIE and Deblurring methods with or without GT reference}, respectively. The best results for each type are highlighted in \textbf{bold}. [Key:  $\uparrow$ ($\downarrow$): Larger (smaller) values leads to better performance, *: using GT mean intensity for illumination adjustment]}
  \label{table_1}
\end{table*}

\subsection{Recursive Phase Modulation}
\label{sec_method_phase_blur}

\begin{figure}[!h]
    \vspace{-1mm}
    \centering
    \setlength{\abovecaptionskip}{1.2mm}
    \centering
    \includegraphics[width=0.9\linewidth]{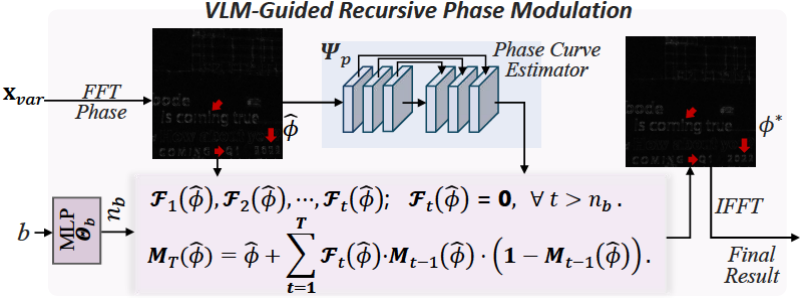}
    \caption{Our VGPM progressively refines the phase representation to mitigate ghosting artifacts introduced by blur.}
    \label{fig_vgpm}
\end{figure}

To address residual motion blur and structural degradation, we propose a VLM-guided recursive phase modulation module (VGPM) applied to the output $\mathbf{x}_{\text{var}}$ of our enhanced VAR backbone. Motivated by observations that blurry inputs often exhibit repeated edge artifacts in the \texttt{FFT} phase domain (see arrows in Fig.~\ref{fig_vgpm}), we employ phase information as a structurally informative representation that is more robust to occlusion ambiguity and illumination noise than spatial-domain features. As shown in Fig.~\ref{fig_vgpm}, we first compute the normalized phase map $\hat{\phi} = (\phi + \pi) / 2\pi \in [0,1]$ ($\phi$: original phase). A recursive enhancement is then performed as:
\begin{equation}
    \begin{split}
    \bm{M}_0(\hat \phi)& = \hat \phi, \\
    \bm{M}_T(\hat \phi) = \hat \phi + \sum_{t=1}^{T} \bm{\mathcal{F}}_t(\hat \phi)&M_{t-1}(\hat \phi)\left(\mathbf{1} - \bm{M}_{t-1}(\hat \phi) \right),
\end{split}
\end{equation}
where $T$ is the total number of modulation steps (set to 8), and $\bm{\mathcal{F}}_t \in [0,1]$ is a phase adjustment map predicted by the estimator $\mathbf{\Psi}_p$, which shares architecture with $\mathbf{\Psi}_I$ in VICM. Based on VLM-based blur assessment $b$,  we further employ a MLP $\mathbf{\Theta}_b$ to adaptively guide the modulation strength . The final enhanced phase $\phi^*$ is obtained by inverting $\bm{M}_T(\hat{\phi})$ back to the original domain and applying an \texttt{IFFT} to produce the restored image $\mathbf{x}_{\text{out}}$.  

\subsection{Optimization}
\label{sec_method_optimization}
We jointly optimize all parameters ($\mathbf{\Theta}_b$, $\mathbf{\Psi}_I$, $\mathbf{\Theta}_v$, $\mathbf{\Psi}_p$, $\lambda$) in a reference-free manner using the following losses. 

\paragraph{Adaptive Exposure Control Loss} We adopt a visibility-aware exposure target, where the base level $E = 0.45$ is dynamically adjusted by $E_d \in [-0.1, 0.1]$ obtained from $\mathbf{\Theta}_v$, and \texttt{Mean} aims to calculate the mean intensity:
\begin{equation}
    \mathcal{L}_{ex} = |\texttt{Mean}(I_{out}) - (E + E_d)|,
\end{equation}

\paragraph{Structural Entropy Loss}  
To promote phase-guided structural fidelity, we reconstruct $\mathbf{S}_\phi^* = |\texttt{IFFT}(e^{j\phi^*})|$ and compute Shannon entropy~\cite{entropy} over its histogram:
\begin{equation}
\mathcal{L}_{en} = -\sum_{i=1}^{B} p_i \log(p_i),
\end{equation}
where $p_i$ is the probability of the $i$-th bin in the normalized histogram of $\mathbf{S}_\phi^*$, $B$ is the total number of bins.

\paragraph{Structural Contrast Loss} 
We improve local structural distinctiveness via negative variance over $N=16$  patches:
\begin{equation}
\mathcal{L}_{con} = -\frac{1}{N} \sum_{k=1}^{N} \sigma^2(\mathbf{S}_{\phi,k}^*).
\end{equation}

\paragraph{Total Variation Loss} To suppress artifacts, we apply a total variation (TV) loss~\cite{TV} on $\mathbf{x}_{out}$:
\begin{equation}
\mathcal{L}_{tv} = \sum_{x, y} \left| I_{x+1, y} - I_{x, y} \right| + \left| I_{x, y+1} - I_{x, y} \right|.
\end{equation}

The overall optimization objective is formulated as:
\begin{equation}
\mathcal{L} = \mathcal{L}_{ex} + \lambda_{en}\mathcal{L}_{en} + \lambda_{con}\mathcal{L}_{con} + \lambda_{tv}\mathcal{L}_{tv},
\end{equation}
where $\lambda_{en}$, $\lambda_{con}$, and $\lambda_{tv}$ are weights of the losses.

\begin{figure*}[!t]
    \centering
    \setlength{\abovecaptionskip}{1mm}
    \centering
    \includegraphics[width=0.87\linewidth]{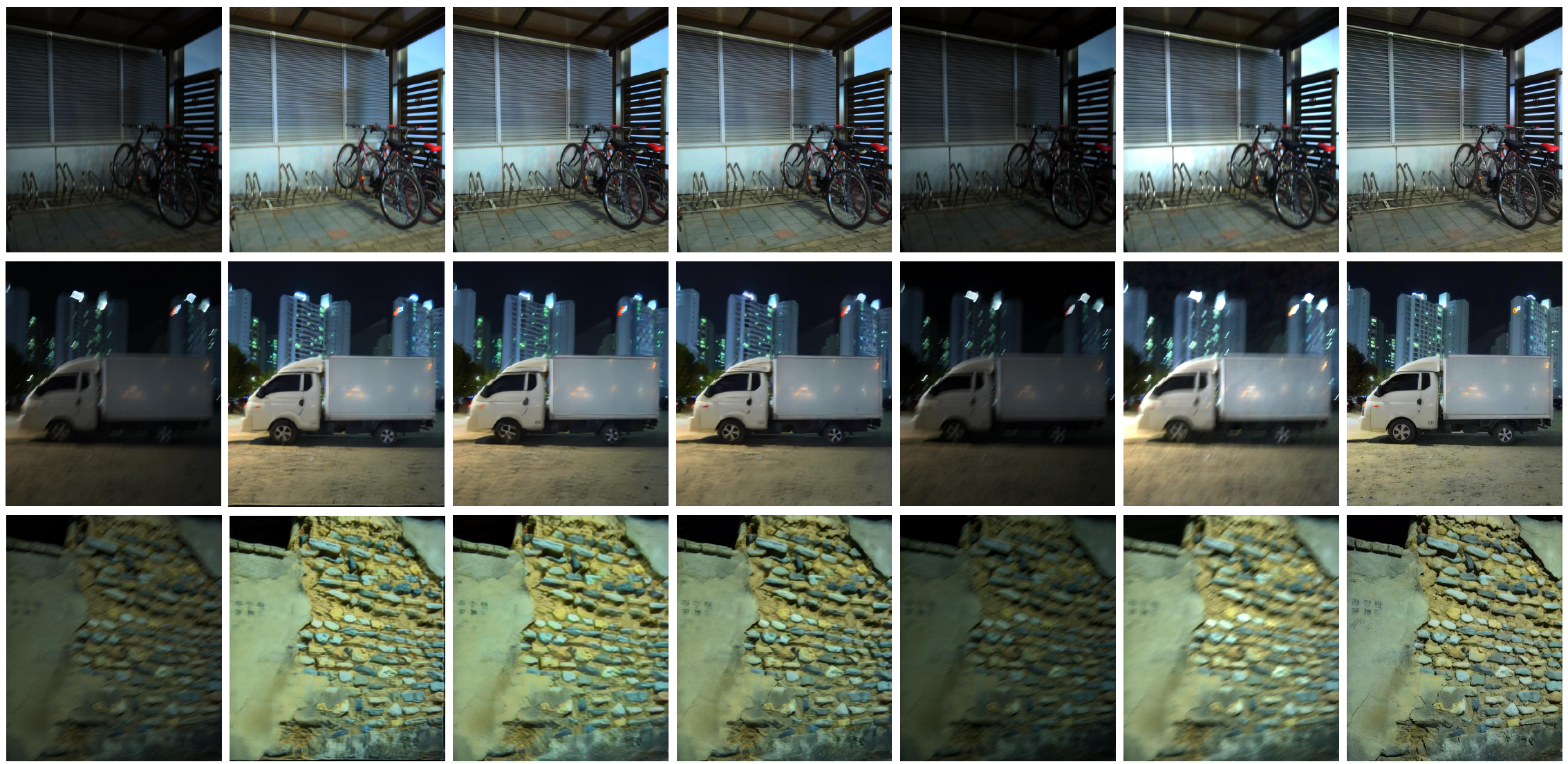}
    \setlength{\parskip}{0mm} 
    \setlength{\baselineskip}{0mm}
    \begin{minipage}{0.8\linewidth}
    \begin{minipage}[b]{0.137\linewidth}
    \centering
    \scriptsize{ LQ}
    \end{minipage}
    \begin{minipage}[b]{0.137\linewidth}
    \centering
    \scriptsize{DarkIR}
    \end{minipage}
    \begin{minipage}[b]{0.137\linewidth}
    \centering
    \scriptsize{LEDNet}
    \end{minipage}
    \begin{minipage}[b]{0.137\linewidth}
    \centering
    \scriptsize{JUDE}
    \end{minipage}
    \begin{minipage}[b]{0.137\linewidth}
    \centering
    \scriptsize{SSFlow}
    \end{minipage}
    \begin{minipage}[b]{0.137\linewidth}
    \centering
    \scriptsize{FourierDiff}
    \end{minipage}
    \begin{minipage}[b]{0.137\linewidth}
    \centering
    \scriptsize{\textbf{Ours}}
    \end{minipage}
\end{minipage}
    \caption{Visual comparisons on the Real-LOLBlur dataset. Our method restores natural illumination and achieves superior deblurring performance with sharper edges and clearer structures, showing strong generalization to complex real-world scenes.}
    \label{fig_realblur}
\end{figure*}


\section{Experiments}
\label{sec_exp}

\subsection{Experiment Settings} \label{sec_imple}
\paragraph{Training Details} Our model is trained on $512 \times 512$ resolution images using the AdamW optimizer for 200 epochs. The initial learning rate is set to $10^{-4}$ and decayed by a factor of 0.5 at epochs 100, 150, 180, and 190. All experiments are conducted on a single NVIDIA 5090 GPU.

\paragraph{Datasets and Metrics} We evaluate our method on the LOLBlur and Real-LOLBlur~\cite{LEDNet} datasets. LOLBlur comprises 12,000 image pairs with diverse illumination and motion blur. We utilize 10,200 low-blur, noisy images from the official training split for optimization, and the remaining 1,800 pairs for both quantitative and qualitative evaluation using reference-based (PSNR, LPIPS~\cite{LPIPS}, and FID~\cite{fid}) and no-reference metrics. Following LEDNet~\cite{LEDNet}, we further assess generalization on Real-LOLBlur dataset, which contains 1,354 unpaired real-world low-light blurry images. Evaluation employs NIQE~\cite{NIQE}, CLIPIQA~\cite{clipiqa}, MUSIQ~\cite{musiq}, and MANIQA~\cite{maniqa}. 


\subsection{Performance on LOLBlur Dataset} 
We compare our method against four baseline categories: \textbf{LLIE}~$\rightarrow$~\textbf{Deblurring}, ~\textbf{Deblurring}~$\rightarrow$~\textbf{LLIE}, supervised joint frameworks, and unsupervised joint frameworks. 

\paragraph{\textbf{LLIE}~$\rightarrow$~\textbf{Deblurring}} LLIE models (SCI~\cite{SCI}, EnlightenGAN~\cite{enlightengan}, and Zero-DCE++~\cite{Zero-DCE++}) are first trained. Their enhanced outputs are subsequently used to train BD\_Noise~\cite{bd_noise} and Blur2Blur~\cite{blur2blur} for deblurring.

\paragraph{\textbf{Deblurring}~$\rightarrow$~\textbf{LLIE}} We begin by training deblurring methods. The outputs are then optimized by LLIE methods.

\paragraph{End-to-End Methods without Reference} Our model, together with SSFlow \cite{SSFlow} and FourierDiff \cite{FourierDiff}, is optimized without using any GT reference images. 

\paragraph{Supervised End-to-End Baselines}  We also incorporate representative supervised models for benchmarking, including  LEDNet~\cite{LEDNet}, LIEDNet~\cite{LIEDNet}, RetinexFormer~\cite{Retinexformer}, JUDE~\cite{JUDE}, and DarkIR~\cite{DarkIR}.


\paragraph{Quantitative and Qualitative Comparisons} As summarized in Tab.~\ref{table_1}, our proposed method significantly outperforms all cascaded pipelines, achieving over 5 dB PSNR gains and superior perceptual quality. Compared to existing unsupervised joint frameworks (SSFlow and FourierDiff), our method exhibits notable improvements across both pixel-level and perceptual metrics. Furthermore, despite optimized without ground-truth supervision, our model achieves performance competitive with fully supervised baselines (\textit{e.g.}, LEDNet and JUDE), particularly excelling in perceptual quality as indicated by the lowest NIQE. 
As illustrated in Fig.~\ref{fig_lolblur}, our method delivers perceptually more faithful reconstructions compared to unsupervised baselines. In particular, it preserves fine-grained texture details and recovers natural illumination more effectively than SSFlow and FourierDiff. While supervised models such as Retinexformer yield enhanced brightness, they often fail to remove motion blur. In contrast, our approach achieves a more balanced restoration with fewer visual artifacts.

\subsection{Performance on Unpaired Real-World Data}

We conduct cross-dataset evaluations on unpaired real-world data. Specifically, we directly apply the model trained on the LOLBlur dataset to unseen samples from Real-LOLBlur dataset. Notably, FourierDiff involves an internal optimization process during inference. Quantitative comparisons and visual results are summarized in Tab.~\ref{table_2} and Fig.~\ref{fig_realblur}. 

\paragraph {Comparison Results} As shown in Tab.~\ref{table_2} and Fig.~\ref{fig_realblur}, our method demonstrates good generalization ability on the Real-LOLBlur dataset, despite being trained solely on LOLBlur without access to paired data. Quantitatively, it achieves the best NIQE, CLIPIQA, and MUSIQ scores among all unsupervised joint methods and even approaches or outperforms several supervised counterparts. Qualitatively, our outputs retain better structural integrity and perceptual fidelity, avoiding over-smoothing or illumination inconsistencies commonly observed in competing baselines.

\begin{table}[t]
\Large
\setlength{\abovecaptionskip}{1mm}
\setlength{\tabcolsep}{1.3pt}
\renewcommand\arraystretch{1.0}
  \centering
  \resizebox{\linewidth}{!}{
  \begin{tabular}{cc|cccc}
    \hline
    \hline
    \multirow{2}{*}{\begin{tabular}{c}
            \textbf{ Methods}
        \end{tabular}} &\multirow{2}{*}{\begin{tabular}{c}
            \textbf{ Type}
        \end{tabular}} &\multicolumn{4}{c}{Metrics} \\
         & &NIQE$\downarrow$ &CLIPIQA$\uparrow$  &MUSIQ$\uparrow$ &MANIQA$\uparrow$ \\ 
    \hline
    \multicolumn{6}{c}{\textbf{Cascaded Methods}} \\ \hline
    EnlightenGAN + BD\_Noise  & L+D  &5.44 &0.157 &\textbf{40.05} &\textbf{0.171} \\
    EnlightenGAN + Blur2Blur & L+D  &5.49 &0.160 &39.24 &0.169  \\
    
    Zero-DCE++ + BD\_Noise & L+D &5.66 &0.161 &25.88 &0.103  \\
    Zero-DCE++ + Blur2Blur & L+D &5.51 &0.167 &27.20 &0.109  \\
    
    SCI + BD\_Noise & L+D  &5.22 &0.181 &34.28 &0.126  \\
    SCI + Blur2Blur  &L+D &\textbf{5.13} &\textbf{0.185} &33.87 &0.129  \\

    BD\_Noise+ EnlightenGAN & D+L &5.58 &0.150 &\textbf{40.36} &\textbf{0.166}    \\
    BD\_Noise + Zero-DCE++ &D+L &5.77 &0.210 &22.41 &0.130   \\
    BD\_Noise + SCI &D+L &5.64 &0.180 &26.37 &0.109   \\
    
    Blur2Blur +EnlightenGAN & D+L &\textbf{5.43} &0.152 &38.96 &0.165   \\
    Blur2Blur + Zero-DCE++ &D+L &5.92 &\textbf{0.215} &22.47 &0.127   \\
    Blur2Blur+ SCI  &D+L &5.52 &0.182 &24.64 &0.102    \\
    \hline
    \multicolumn{6}{c}{\textbf{Joint LLIE and Deblurring Methods}} \\ \hline
    LEDNet & w R &5.07  &0.256   &49.46  &\textbf{0.228}  \\
    Retinexformer  & w R &5.69 &0.208 &40.47 &0.173 \\
    JUDE  & w R &4.92  &0.236  &\textbf{50.29} &0.223  \\
    DarkIR-M  & w R  &4.97 &0.252 &48.31 &0.209 \\
    DarkIR-L & w R  & \textbf{4.90} &\textbf{0.262} &48.72 &0.216 \\
    
    SSFlow  & w/o R &5.94 &0.190 &30.93 &0.148   \\
    FourierDiff  & w/o R   &5.59 &0.187 &32.01 &0.122   \\
    \textbf{Ours}  & w/o R  &\textbf{5.16} &\textbf{0.226} &\textbf{47.53} &\textbf{0.223}   \\

  \hline
  \hline
  \end{tabular}
  }
  \caption{Quantitative comparisons on Real-LOLBlur Dataset. Our method achieves superior perceptual quality compared to unsupervised baselines.}
  \label{table_2}
\end{table}

\section{Conclusion}
\label{conclusion}

In this work, we introduce \textbf{VAR-LIDE}, a fully unsupervised generative framework for joint LLIE and deblurring. By leveraging the autoregressive modeling capacity of VAR backbone and the perceptual guidance from VLMs, we design a VLM-informed conditioning mechanism that achieve adaptive illumination enhancement. To further strengthen structural fidelity under blur, we enhance the positional modeling of the VAR backbone via input-adaptive spatial-frequency RoPE. Additionally, a recursive phase-domain modulation module is developed to suppress blur-induced edge artifacts, with guidance from blur-aware VLM assessments. Extensive experiments confirm that VAR-LIDE achieves impressive quantitative and perceptual performance on several benchmarks.

\bibliography{aaai2026}

@inproceedings{Zero-DCE,
  author = {Guo, Chunle and Li, Chongyi and Guo, Jichang and Loy, Chen Change and Hou, Junhui and Kwong, Sam and Cong, Runmin},
  year = {2020},
  title = {Zero-reference deep curve estimation for low-light image enhancement},
  booktitle = {Proceedings of the IEEE/CVF conference on computer vision and pattern recognition},
  pages = {1780-1789},
}

@article{Zero-DCE++,
  author = {Li, Chongyi and Guo, Chunle and Loy, Chen Change},
  journal = {IEEE Transactions on Pattern Analysis and Machine Intelligence}, 
  title = {Learning to Enhance Low-Light Image via Zero-Reference Deep Curve Estimation}, 
  year = {2022},
  volume = {44},
  number = {8},
  pages = {4225-4238},
  doi = {10.1109/TPAMI.2021.3063604},
}

@article{enlightengan,
  author = {Jiang, Yifan and Gong, Xinyu and Liu, Ding and Cheng, Yu and Fang, Chen and Shen, Xiaohui and Yang, Jianchao and Zhou, Pan and Wang, Zhangyang},
  journal = {IEEE Transactions on Image Processing},
  title = {EnlightenGAN: Deep Light Enhancement Without Paired Supervision},
  year = {2021},
  volume = {30},
  number = {},
  pages = {2340-2349},
}

@article{retinex,
  author = {Jobson, D.J. and Rahman, Z. and Woodell, G.A.},
  journal = {IEEE Transactions on Image Processing},
  title = {A multiscale retinex for bridging the gap between color images and the human observation of scenes},
  year = {1997},
  volume = {6},
  number = {7},
  pages = {965-976},
  doi = {10.1109/83.597272},
}

@inproceedings{GRL,
  author = {Li, Yawei and Fan, Yuchen and Xiang, Xiaoyu and Demandolx, Denis and Ranjan, Rakesh and Timofte, Radu and Van Gool, Luc},
  year = {2023},
  title = {Efficient and explicit modelling of image hierarchies for image restoration},
  booktitle = {Proceedings of the IEEE/CVF conference on computer vision and pattern recognition},
  pages = {18278-18289},
}

@inproceedings{LEDNet,
  author = {Zhou, Shangchen and Li, Chongyi and Change Loy, Chen},
  year = {2022},
  title = {Lednet: Joint low-light enhancement and deblurring in the dark},
  booktitle = {European conference on computer vision},
  pages = {573-589},
  publisher = {Springer},
}

@inproceedings{SCI,
  author = {Ma, Long and Ma, Tengyu and Liu, Risheng and Fan, Xin and Luo, Zhongxuan},
  year = {2022},
  title = {Toward Fast, Flexible, and Robust Low-Light Image Enhancement},
  booktitle = {Proceedings of the IEEE/CVF Conference on Computer Vision and Pattern Recognition},
  pages = {5637-5646},
}

@inproceedings{DarkIR,
  author = {Feijoo, Daniel and Benito, Juan C and Garcia, Alvaro and Conde, Marcos V},
  year = {2025},
  title = {Darkir: Robust low-light image restoration},
  booktitle = {Proceedings of the Computer Vision and Pattern Recognition Conference},
  pages = {10879-10889},
}

@inproceedings{Retinexformer,
  author = {Cai, Yuanhao and Bian, Hao and Lin, Jing and Wang, Haoqian and Timofte, Radu and Zhang, Yulun},
  year = {2023},
  title = {Retinexformer: One-stage retinex-based transformer for low-light image enhancement},
  booktitle = {Proceedings of the IEEE/CVF international conference on computer vision},
  pages = {12504-12513},
}

@inproceedings{FourierDiff,
  author = {Lv, Xiaoqian and Zhang, Shengping and Wang, Chenyang and Zheng, Yichen and Zhong, Bineng and Li, Chongyi and Nie, Liqiang},
  year = {2024},
  title = {Fourier priors-guided diffusion for zero-shot joint low-light enhancement and deblurring},
  booktitle = {Proceedings of the IEEE/CVF Conference on Computer Vision and Pattern Recognition},
  pages = {25378-25388},
}

@article{HistogramEqualization,
  author = {Stephen M. Pizer and E. Philip Amburn and John D. Austin and Robert Cromartie and Ari Geselowitz and Trey Greer and Bart {ter Haar Romeny} and John B. Zimmerman and Karel Zuiderveld},
  title = {Adaptive histogram equalization and its variations},
  volume = {39},
  number = {3},
  pages = {355-368},
  year = {1987},
  doi = {https://doi.org/10.1016/S0734-189X(87)80186-X},
  publisher = {},
  issn = {0734-189X},
  URL = {https://www.sciencedirect.com/science/article/pii/S0734189X8780186X},
  eprint = {},
  journal = {Computer Vision, Graphics, and Image Processing},
}

@article{NaturalnessPreservedEnhancement,
  author = {Wang, Shuhang and Zheng, Jin and Hu, Hai-Miao and Li, Bo},
  journal = {IEEE Transactions on Image Processing}, 
  title = {Naturalness Preserved Enhancement Algorithm for Non-Uniform Illumination Images}, 
  year = {2013},
  volume = {22},
  number = {9},
  pages = {3538-3548},
  doi = {10.1109/TIP.2013.2261309}
}

@inproceedings{RealisticDegradations,
  author = {Anger, Jérémy and Facciolo, Gabriele and Delbracio, Mauricio},
  booktitle = {2018 25th IEEE International Conference on Image Processing (ICIP)}, 
  title = {Modeling Realistic Degradations in Non-Blind Deconvolution}, 
  year = {2018},
  volume = {},
  number = {},
  pages = {978-982},
  doi = {10.1109/ICIP.2018.8451115},
}

@article{BlindImageDeconvolution,
  author = {Kundur, D. and Hatzinakos, D.},
  journal = {IEEE Signal Processing Magazine}, 
  title = {Blind image deconvolution}, 
  year = {1996},
  volume = {13},
  number = {3},
  pages = {43-64},
  doi = {10.1109/79.489268},
}

@article{DWDN,
  author = {Dong, Jiangxin and Roth, Stefan and Schiele, Bernt},
  journal = {IEEE Transactions on Pattern Analysis and Machine Intelligence}, 
  title = {DWDN: Deep Wiener Deconvolution Network for Non-Blind Image Deblurring}, 
  year = {2022},
  volume = {44},
  number = {12},
  pages = {9960-9976},
  doi = {10.1109/TPAMI.2021.3138787},
}

@inproceedings{MambaIR,
  author = {Guo, Hang and Li, Jinmin and Dai, Tao and Ouyang, Zhihao and Ren, Xudong and Xia, Shu-Tao},
  year = {2024},
  title = {Mambair: A simple baseline for image restoration with state-space model},
  booktitle = {European conference on computer vision},
  pages = {222-241},
  publisher = {Springer}
}

@inproceedings{VarFormer,
  author = {Wang, Siyang and Zheng, Naishan and Huang, Jie and Zhao, Feng},
  year = {2025},
  title = {Navigating Image Restoration with VAR's Distribution Alignment Prior},
  booktitle = {Proceedings of the Computer Vision and Pattern Recognition Conference},
  pages = {7559-7569},
}

@misc{RestoreVAR,
  title = {RestoreVAR: Visual Autoregressive Generation for All-in-One Image Restoration}, 
  author = {Sudarshan Rajagopalan and Kartik Narayan and Vishal M. Patel},
  year = {2025},
  eprint = {2505.18047},
  archivePrefix = {arXiv},
  primaryClass = {cs.CV},
}

@inproceedings{blur2blur,
  author = {Pham, Bang-Dang and Tran, Phong and Tran, Anh and Pham, Cuong and Nguyen, Rang and Hoai, Minh},
  year = {2024},
  title = {Blur2blur: Blur conversion for unsupervised image deblurring on unknown domains},
  booktitle = {Proceedings of the IEEE/CVF Conference on Computer Vision and Pattern Recognition},
  pages = {2804-2813},
}

@article{LIEDNet,
  author = {Liu, Mingyu and Cui, Yuning and Ren, Wenqi and Zhou, Juxiang and Knoll, Alois C.},
  journal = {IEEE Transactions on Circuits and Systems for Video Technology}, 
  title = {LIEDNet: A Lightweight Network for Low-Light Enhancement and Deblurring}, 
  year = {2025},
  volume = {35},
  number = {7},
  pages = {6602-6615},
  doi = {10.1109/TCSVT.2025.3541429},
}

@inproceedings{CoDe,
  author = {Chen, Zigeng and Ma, Xinyin and Fang, Gongfan and Wang, Xinchao},
  year = {2025},
  title = {Collaborative decoding makes visual auto-regressive modeling efficient},
  booktitle = {Proceedings of the Computer Vision and Pattern Recognition Conference},
  pages = {23334-23344},
}

@inproceedings{Infinity,
  author = {Han, Jian and Liu, Jinlai and Jiang, Yi and Yan, Bin and Zhang, Yuqi and Yuan, Zehuan and Peng, Bingyue and Liu, Xiaobing},
  year = {2025},
  title = {Infinity: Scaling bitwise autoregressive modeling for high-resolution image synthesis},
  booktitle = {Proceedings of the Computer Vision and Pattern Recognition Conference},
  pages = {15733-15744},
}

@article{SSFlow,
  title = {Self-supervised normalizing flow for jointing low-light enhancement and deblurring},
  author = {Li, Lingyan and Zhu, Chunzi and Chen, Jiale and Shi, Baoshun and Lian, Qiusheng},
  journal = {Circuits, Systems, and Signal Processing},
  volume = {43},
  number = {9},
  pages = {5727-5748},
  year = {2024},
  publisher = {Springer},
}

@inproceedings{JUDE,
  author = {Vo, Tu and Park, Chan Y.},
  booktitle = {2025 IEEE/CVF Winter Conference on Applications of Computer Vision (WACV)}, 
  title = {Deep Joint Unrolling for Deblurring and Low-Light Image Enhancement (JUDE)}, 
  year = {2025},
  volume = {},
  number = {},
  pages = {2696-2705},
  doi = {10.1109/WACV61041.2025.00267},
}

@inproceedings{varsr,
  title={Visual Autoregressive Modeling for Image Super-Resolution},
  author={Qu, Yunpeng and Yuan, Kun and Hao, Jinhua and Zhao, Kai and Xie, Qizhi and Sun, Ming and Zhou, Chao},
  booktitle={Proceedings of the 32nd International Conference on Machine Learning (ICML)},
  year={2025}
}

@inproceedings{var,
  title={Visual Autoregressive Modeling: Scalable Image Generation via Next-Scale Prediction},
  author={Keyu Tian and Yi Jiang and Zehuan Yuan and Bingyue Peng and Liwei Wang},
  booktitle={Advances in Neural Information Processing Systems (NeurIPS)},
  year={2024}
}

@inproceedings{vqvae,
  title={Neural discrete representation learning},
  author={Aaron van den Oord and Oriol Vinyals and Koray Kavukcuoglu},
  booktitle={Advances in Neural Information Processing Systems (NeurIPS)},
  year={2017}
}

@inproceedings{gppllie,
  title={Low-light image enhancement via generative perceptual priors},
  author={Zhou, Han and Dong, Wei and Liu, Xiaohong and Zhang, Yulun and Zhai, Guangtao and Chen, Jun},
  booktitle={Proceedings of the AAAI Conference on Artificial Intelligence},
  year={2025}
}

@inproceedings{clipiqa,
  title={Exploring clip for assessing the look and feel of images},
  author={Wang, Jianyi and Chan, Kelvin CK and Loy, Chen Change},
  booktitle={Proceedings of the AAAI conference on artificial intelligence},
  year={2023}
}

@inproceedings{musiq,
  title={Musiq: Multi-scale image quality transformer},
  author={Ke, Junjie and Wang, Qifei and Wang, Yilin and Milanfar, Peyman and Yang, Feng},
  booktitle={Proceedings of the IEEE/CVF international conference on computer vision},
  year={2021}
}

@inproceedings{maniqa,
  title={Maniqa: Multi-dimension attention network for no-reference image quality assessment},
  author={Yang, Sidi and Wu, Tianhe and Shi, Shuwei and Lao, Shanshan and Gong, Yuan and Cao, Mingdeng and Wang, Jiahao and Yang, Yujiu},
  booktitle={Proceedings of the IEEE/CVF conference on computer vision and pattern recognition},
  year={2022}
}

@article{NIQE,
  title={Making a “completely blind” image quality analyzer},
  author={Mittal, Anish and Soundararajan, Rajiv and Bovik, Alan C},
  journal={IEEE Signal processing letters},
  year={2012}
}

@inproceedings{LPIPS,
author = {Richard Zhang and Phillip Isola and Alexei A Efros and Eli Shechtman and Oliver Wang},
title = {The unreasonable effectiveness of deep features as a perceptual metric},
booktitle = {Proceedings of the IEEE/CVF conference on computer vision and pattern recognition},
year = {2018}
}

@inproceedings{fid,
    author={Martin Heusel and Hubert Ramsauer and Thomas Unterthiner and Bernhard Nessler and Sepp Hochreiter},
    title = {Gans trained by a two time-scale update rule converge to a local nash equilibrium},
    booktitle = {Advances in neural information processing systems},
    year = {2017}
}

@inproceedings{glare,
  title={Glare: Low light image enhancement via generative latent feature based codebook retrieval},
  author={Zhou, Han and Dong, Wei and Liu, Xiaohong and Liu, Shuaicheng and Min, Xiongkuo and Zhai, Guangtao and Chen, Jun},
  booktitle={European Conference on Computer Vision},
  year={2024},
}

@inproceedings{bd_noise,
  title={Blind Image Deblurring with Noise-Robust Kernel Estimation},
  author={Lee, Chanseok and Kim, Jeongsol and Lee, Seungmin and Jung, Jaehwang and Cho, Yunje and Kim, Taejoong and Jo, Taeyong and Lee, Myungjun and Jang, Mooseok},
  booktitle={European Conference on Computer Vision},
  pages={188--204},
  year={2024},
  organization={Springer}
}

@article{entropy,
  title={Divergence measures based on the Shannon entropy},
  author={Lin, Jianhua},
  journal={IEEE Transactions on Information theory},
  year={2002},
  publisher={IEEE}
}

@article{TV,
  title={An iterative regularization method for total variation-based image restoration},
  author={Stanley Osher and Martin Burger and Donald Goldfarb and Jinjun Xu and Wotao Yin},
  journal={Multiscale Modeling \& Simulation},
  year={2005}
}

@article{ecmamba,
  title={Ecmamba: Consolidating selective state space model with retinex guidance for efficient multiple exposure correction},
  author={Dong, Wei and Zhou, Han and Zhang, Yulun and Liu, Xiaohong and Chen, Jun},
  journal={Advances in Neural Information Processing Systems},
  year={2024}
}

@inproceedings{zhou2023breaking,
  title={Breaking through the haze: An advanced non-homogeneous dehazing method based on fast fourier convolution and convnext},
  author={Zhou, Han and Dong, Wei and Liu, Yangyi and Chen, Jun},
  booktitle={Proceedings of the IEEE/CVF Conference on Computer Vision and Pattern Recognition},
  year={2023}
}

@inproceedings{litags,
  title={LITA-GS: Illumination-Agnostic Novel View Synthesis via Reference-Free 3D Gaussian Splatting and Physical Priors},
  author={Zhou, Han and Dong, Wei and Chen, Jun},
  booktitle={Proceedings of the Computer Vision and Pattern Recognition Conference},
  year={2025}
}

@inproceedings{dehazedct,
  title={Dehazedct: Towards effective non-homogeneous dehazing via deformable convolutional transformer},
  author={Dong, Wei and Zhou, Han and Wang, Ruiyi and Liu, Xiaohong and Zhai, Guangtao and Chen, Jun},
  booktitle={Proceedings of the IEEE/CVF Conference on Computer Vision and Pattern Recognition},
  year={2024}
}

@inproceedings{dong2024shadowrefiner,
  title={ShadowRefiner: Towards mask-free shadow removal via fast fourier transformer},
  author={Dong, Wei and Zhou, Han and Tian, Yuqiong and Sun, Jingke and Liu, Xiaohong and Zhai, Guangtao and Chen, Jun},
  booktitle={Proceedings of the IEEE/CVF Conference on Computer Vision and Pattern Recognition},
  year={2024}
}

@inproceedings{sgllie,
  title={Towards Scale-Aware Low-Light Enhancement via Structure-Guided Transformer Design},
  author={Dong, Wei and Min, Yan and Zhou, Han and Chen, Jun},
  booktitle={Proceedings of the Computer Vision and Pattern Recognition Conference},
  year={2025}
}

@inproceedings{rehit,
  title={Retinex-guided histogram transformer for mask-free shadow removal},
  author={Dong, Wei and Zhou, Han and Mousavi, Seyed Amirreza and Chen, Jun},
  booktitle={Proceedings of the Computer Vision and Pattern Recognition Conference},
  year={2025}
}

@article{sclera,
  title={Sclera recognition based on efficient sclera segmentation and significant vessel matching},
  author={Xu, Dong and Dong, Wei and Zhou, Han},
  journal={The Computer Journal},
  year={2022}
}

\end{document}